\documentclass[sigconf]{acmart}
\AtBeginDocument{%
  \providecommand\BibTeX{{%
    \normalfont B\kern-0.5em{\scshape i\kern-0.25em b}\kern-0.8em\TeX}}}

\setcopyright{acmcopyright}
\copyrightyear{2023}
\acmYear{2023}
\acmDOI{XXXXXXX.XXXXXXX}
\acmConference[MM '23]{Proceedings of the 31th ACM International Conference on Multimedia}{October 29 – November 3, 2023}{Ottawa, Canada}
\acmBooktitle{Proceedings of the 31th ACM International Conference on Multimedia (MM '23), October 29 – November 3, 2023, Ottawa, Canada}
\acmPrice{15.00}
\acmISBN{978-1-4503-XXXX-X/23/10}

\usepackage{booktabs, multirow, placeins, float, graphicx, indentfirst, mathrsfs, amsmath, tabularx, adjustbox}
\usepackage{array}
\usepackage{booktabs, pifont, subcaption, caption}

\begin{document}

%%%%%%%%%%%%%%%%%%%%%%%%%%%%%%%%%%%%%% title %%%%%%%%%%%%%%%%%%%%%%%%%%%%%%%%%%%%%%
\title{MH-DETR: Video Moment and Highlight Detection with Cross-modal Transformer}

%%%%%%%%%%%%%%%%%%%%%%%%%%%%%%%%%%%%%% author %%%%%%%%%%%%%%%%%%%%%%%%%%%%%%%%%%%%%%


\author{Yifang Xu}
\affiliation{%
  \institution{Nanjing University}
  \city{Nanjing}
  \state{Jiangsu}
  \country{China}
}
\email{xyf@smail.nju.edu.cn}

\author{Yunzhuo Sun}
\affiliation{%
  \institution{Hubei Normal University}
  \city{Huangshi}
  \state{Hubei}
  \country{China}
}
\email{sunyunzhuo98@outlook.com}

\author{Yang Li}
\affiliation{%
  \institution{Nanjing University}
  \city{Nanjing}
  \state{Jiangsu}
  \country{China}
}
\email{yogo@nju.edu.cn}

% \author{Zien Xie}
% \affiliation{%
%   \institution{Nanjing University}
%   \city{Nanjing}
%   \state{Jiangsu}
%   \country{China}
% }
% \email{TODO}

\author{Yilei Shi}
\affiliation{%
  \institution{MedAI Technology (Wuxi) Co., Ltd.}
  \city{Wuxi}
  \state{Jiangsu}
  \country{China}
}
\email{yilei.shi@medimagingai.com}

\author{Xiaoxiang Zhu}
\affiliation{%
  \institution{Technical University of Munich}
  % \city{Wuxi}
  \state{Munich}
  \country{Germany}
}
\email{xiaoxiang.zhu@tum.de}

\author{Sidan Du}
\authornote{Corresponding author.}
\affiliation{%
  \institution{Nanjing University}
  \city{Nanjing}
  \state{Jiangsu}
  \country{China}
}
\email{coff128@nju.edu.cn}

% \affiliation{%

% \affiliation{%

\renewcommand{\shortauthors}{Trovato and Tobin, et al.}

%%%%%%%%%%%%%%%%%%%%%%%%%%%%%%%%%%%%%% abstract, concepts, keywords %%%%%%%%%%%%%%%%%%%%%%%%%%%%%%%%%%%%%%

\begin{abstract}

% With the increasing demand for video understanding, video moment and highlight detection (MHD) has emerged as a critical research topic. MHD aims to localize all moments and predict clip-wise saliency scores simultaneously. Despite progress made by existing DETR-based methods, we observe that these methods coarsely fuse features from different modalities, which weakens the temporal intra-modal context and results in insufficient cross-modal interaction. To address this issue, we propose \textbf{MH-DETR} (\textbf{M}oment and \textbf{H}ighlight \textbf{DE}tection \textbf{TR}ansformer) tailored for MHD. Specifically, we introduce a simple yet efficient pooling operator within the uni-modal encoder to capture global intra-modal context. Moreover, to obtain temporally aligned cross-modal features, we design a plug-and-play cross-modal interaction module between the encoder and decoder, seamlessly integrating visual and textual features. Comprehensive experiments on QVHighlights, Charades-STA, Activity-Net, and TVSum datasets show that MH-DETR outperforms existing state-of-the-art methods, demonstrating its effectiveness and superiority. Our code is available at -.

With the increasing demand for video understanding, video moment and highlight detection (MHD) has emerged as a critical research topic. MHD aims to localize all moments and predict clip-wise saliency scores simultaneously. Despite progress made by existing DETR-based methods, we observe that these methods coarsely fuse features from different modalities, which weakens the temporal intra-modal context and results in insufficient cross-modal interaction. To address this issue, we propose \textbf{MH-DETR} (\textbf{M}oment and \textbf{H}ighlight \textbf{DE}tection \textbf{TR}ansformer) tailored for MHD. Specifically, we introduce a simple yet efficient pooling operator within the uni-modal encoder to capture global intra-modal context. Moreover, to obtain temporally aligned cross-modal features, we design a plug-and-play cross-modal interaction module between the encoder and decoder, seamlessly integrating visual and textual features. Comprehensive experiments on QVHighlights, Charades-STA, Activity-Net, and TVSum datasets show that MH-DETR outperforms existing state-of-the-art methods, demonstrating its effectiveness and superiority. Our code is available at \url{https://github.com/YoucanBaby/MH-DETR}.

\end{abstract}

% In summeration, our main contributions are as follows：
%     1. We propose a novel model MH-DETR, which is a encoder-decoder architecture, where the uni-modal encoder introduce a simple pooling operator as token mixer component to capture global intra-modal context.
%     2. To efficiently fuse visual and textual features, we design 一个即插即用 the cross-modal interaction module bewteen encoder and decoder, which can obtain temporally aligend cross-modal features.
%     3. Comprehensive experiments show that MH-DETR outperforms state-of-the-art methods on QVHighlights, Charades-STA, Activity-Net, TVSum datasets.



\maketitle

%%%%%%%%%%%%%%%%%%%%%%%%%%%%%%%%%%%%%% 正文 %%%%%%%%%%%%%%%%%%%%%%%%%%%%%%%%%%%%%%

%%%%%%%%%%%%%%%%%%%%%%%%%%%%%%%%%%%%%%%%%%%%%%%%%%%%%%%%%%%%%%%%%%%%%%%%%%%%%%%%%%%%%%%%%%
%%%%%%%%%%%%%%%%%%%%%%%%%%%%%%%%%%%%%% Introduction %%%%%%%%%%%%%%%%%%%%%%%%%%%%%%%%%%%%%%
%%%%%%%%%%%%%%%%%%%%%%%%%%%%%%%%%%%%%%%%%%%%%%%%%%%%%%%%%%%%%%%%%%%%%%%%%%%%%%%%%%%%%%%%%%

\begin{figure}[t!]
  \centering
  \includegraphics[width=\linewidth]{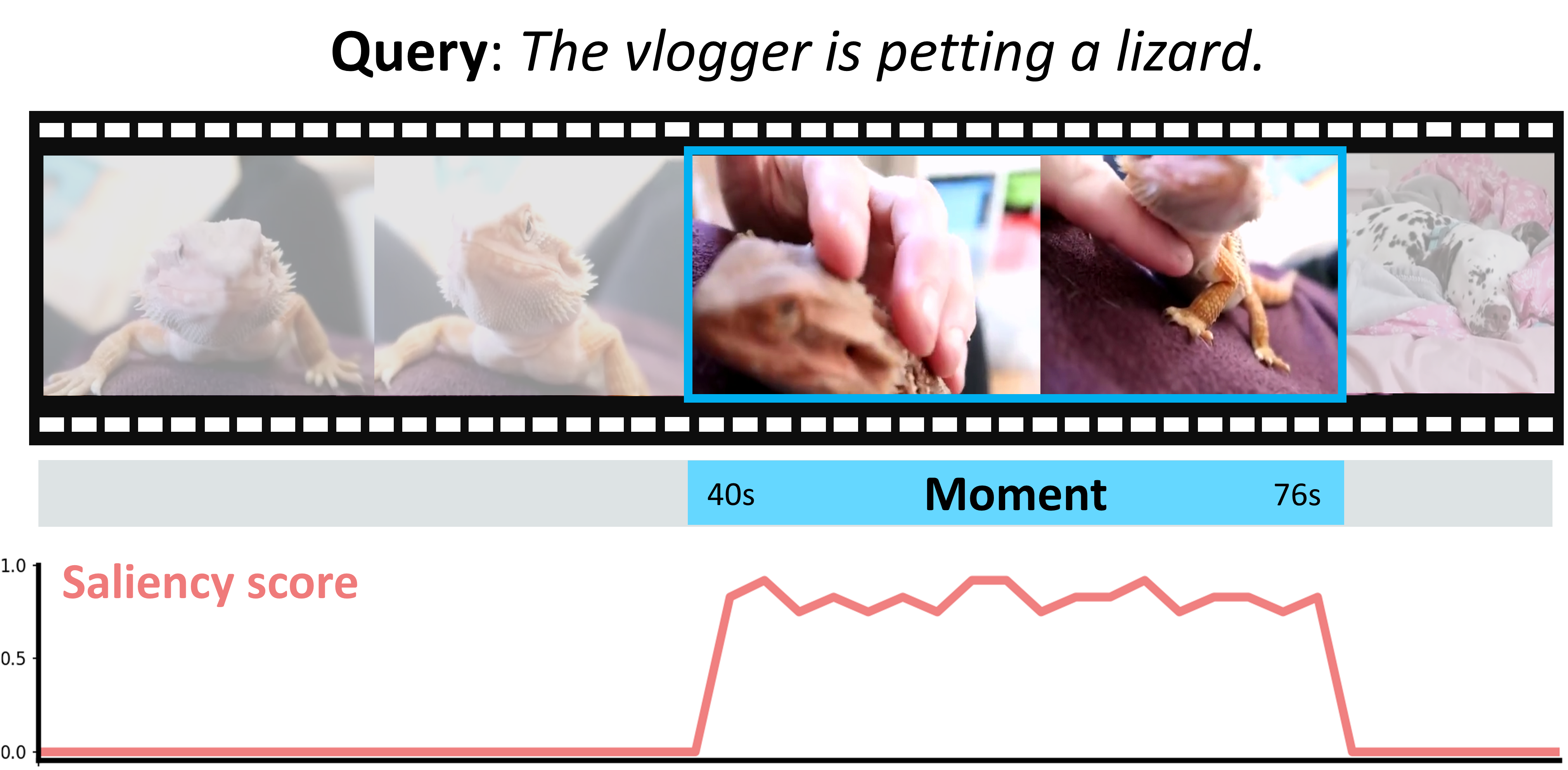}
  \caption{
    An illustrative example of the MHD (video moment and highlight detection) task. Given a video and a natural language query, MHD aims to localize all the moments and predict clip-wise saliency scores simultaneously.
  }
  \label{Fig:1}
\end{figure}

\section{Introduction}

With the rapid development of video creation technologies and internet, a vast number of videos are produced and uploaded to online video platforms every day. Video provides more complex activities and richer semantic information compared to image and text, which results in an increased cost of video understanding \cite{Survey-TSGV-2022}. Consequently, tasks about efficiently searching and browsing video receive increasing attention, including video summarization \cite{Suvery-Video-Summarization-2021, 3DST-UNet-RL-2022}, video captioning \cite{PDVC-2021, MV-GPT-2022, HM-Net-2022, SWINBERT-2022}, temporal action localization \cite{LSBF-2021, FGCL-TAL-2022, PAL-2022}, video moment retrieval (MR) \cite{CTRL-2017, TGN-2018} and video highlight detection (HD) \cite{LIM-S-2019, LPD-2022}. MR aims to retrieve the most relevant moments in a video based on a textual query. HD aims to obtain the saliency scores for each clip in video. Recently, Moment-DETR \cite{MomentDETR-2021} discover the close relationship between these two tasks, and propose a QVHighlights dataset for simultaneously detecting moments and highlights in videos. An illustrative example of MHD (video moment and highlight detection) is shown in Figure \ref{Fig:1}. MHD is even more challenging as it requires not only capturing temporal intra-modal context but also aligned cross-modal interaction of visual and textual features.

Moment-DETR, a baseline model similar to DETR \cite{DETR}, employs an end-to-end transformer encoder-decoder architecture without any time-consuming pre-processing or post-processing steps, such as proposal generation or non-maximum suppression. However, this model naively concatenates visual and textual features and feeds them into transformer encoder. This simplistic approach weakens the global context of uni-modal features and impairs the correlations between cross-modal features. UMT \cite{UMT-2022} proposes a unified architecture to handle diverse inputs, such as video and audio. However, UMT removes the critical components in Moment-DETR, namely the moment decoder and bipartite matching, which leads to inferior performance on moment retrieval.

To tackle the above issues, we propose a novel model called \textbf{MH-DETR} (\textbf{M}oment and \textbf{H}ighlight \textbf{DE}tection \textbf{TR}ansformer), which is based on the end-to-end encoder-decoder architecture. Firstly, in order to model global intra-modal context, we introduce a simple yet efficient pooling operator within the uni-modal encoder, serving as a token mixer component. Next, to obtain temporally aligned cross-modal features, we design a plug-and-play cross-modal interaction module between the encoder and decoder, which can effectively fuse visual and textual features. Specifically, our goal is to emphasize the aspects of the visual content most relevant to the textual semantics. 

To demonstrate the effectiveness of the proposed MH-DETR, we conduct experiments on the QVHighlights dataset and other well-known public datasets for moment retrieval (Charades-STA and Activity-Net) and highlight detection (TVSum). Our extensive experiments show that MH-DETR outperforms state-of-the-art methods on these datasets. All experiments are trained from scratch, utilizing only visual and textual features, without the requirement for additional pre-training. Furthermore, we conduct thorough ablation studies to evaluate essential modules and deliver insightful observations.

%%%%%%%%%%%%%%%%%%%%%%%%%%%%%%%%%%%%%%%%%%%%%%%%%%%%%%%%%%%%%%%%%%%%%%%%%%%%%%%%%%%%%%%%%%%
%%%%%%%%%%%%%%%%%%%%%%%%%%%%%%%%%%%%%% Related works %%%%%%%%%%%%%%%%%%%%%%%%%%%%%%%%%%%%%%
%%%%%%%%%%%%%%%%%%%%%%%%%%%%%%%%%%%%%%%%%%%%%%%%%%%%%%%%%%%%%%%%%%%%%%%%%%%%%%%%%%%%%%%%%%%
\section{Related works}

%%%%%%%%%%%%%%%%%%%%%%%%%%%%%%%%%%%%%% Video Moment and Highlight Detection %%%%%%%%%%%%%%%%%%%%%%%%%%%%%%%%%%%%%%

\textbf{Moment Retrieval.}
Moment retrieval (MR) involves localizing relevant video moments based on a textual query. Existing works mainly utilize proposal-based \cite{MCN-2017, CTRL-2017, 2D-TAN, VLGNet-2021} or proposal-free \cite{GTR-2021, HLGT-2022} methods. The proposal-based methods require additional pre-processing and post-processing steps. Specifically, pre-processing leads to computational redundancy, since it densely samples visual features using sliding windows \cite{MCN-2017, CTRL-2017, MLLC-2018} or anchors \cite{TGN-2018, 2D-TAN, VLGNet-2021} to generate proposals. The time-consuming post-processing step involves non-maximum suppression (NMS). In contrast, proposal-free methods based on end-to-end architecture directly predict start and end clips on moment features. GTR \cite{GTR-2021} is the first DETR-based \cite{DETR} framework for moment retrieval. HLGT \cite{HLGT-2022} utilizes hierarchically local and global information based on GTR. Recently, to utilize cross-modal learning, some works \cite{DRFT-2021, LEORN-2022, DORi-2021} introduce additional features from different modalities, such as depth and optical flow \cite{DRFT-2021}, and object features \cite{LEORN-2022, DORi-2021}.

\vspace{2pt}
\textbf{Highlight Detection.}
Highlight detection (HD) aims to capture highlight clips in the input video, which means predicting clip-wise saliency scores. In contrast to MR, HD methods do not take text as input. As a result, cross-modal methods \cite{VHGNN-2020, LPD-2022, JVA-2021} in HD mainly focus on audio features and different formulations of visual features. VH-GNN \cite{VHGNN-2020} introduces object region proposals and features and uses graph neural networks to model object relationships. LPD \cite{LPD-2022} employs visual saliency features to capture pixel-level differences in the individual video, while SA \cite{JVA-2021} integrates audio and visual features using a bimodal attention mechanism.

\vspace{2pt}
\textbf{Moment and Highlight Detection.} Both MR and HD tasks necessitate learning the correlation between video and textual queries. Historically, these two tasks have been investigated independently. To bridge this gap, Moment-DETR introduced the QVHighlights dataset, targeting the simultaneous detection of moments and highlights in videos (MHD). Moment-DETR, serving as a baseline, employs a transformer encoder-decoder architecture, coarsely fusing visual and textual features within the encoder. Building on Moment-DETR, UMT \cite{UMT-2022} proposed a unified architecture for processing input video and audio. However, it removed the moment decoder and bipartite matching, resulting in inferior performance on MR. In contrast to these works, we present a novel architecture that decouples uni-modal encoding and cross-modal interaction, effectively addressing this challenge.

\vspace{2pt}
\textbf{Cross-modal Learning.} In cross-modal learning, the two most critical aspects are fusion and alignment \cite{Survey-MML-2022}. TERAN \cite{TERAN-2021} introduces a transformer encoder reasoning network for cross-modal retrieval tasks, performing image-sentence matching based on word-region alignments. HGSPN \cite{HGSPN-2019} presents a hierarchical graph semantic pooling network to model hierarchical semantic-level interactions in the cross-modal community question-answering task. AVS \cite{AVSpatialAlignment-2020} employs a cross-attention mechanism with the contrastive loss for the audio-visual temporal synchronization task, aiming to model spatially aligned multi-view video and audio clips. Unlike these tasks, cross-modal interaction in the MHD task focuses on temporal alignment. Therefore, we design a plug-and-play cross-modal interaction module between the encoder and decoder to fuse visual and textual features, resulting in temporally aligned cross-modal features.

\vspace{2pt}
\textbf{General Uni-modal Encoder.} Transformer \cite{Transformer} is initially introduced for machine translation task and rapidly becomes the primary method for various NLP tasks. A series of subsequent works \cite{BERT-2019, TensorizedTransformer-2019} build upon transformer to achieve further improvements. Owing to the success of the transformer in NLP, many studies apply it to CV tasks \cite{ViT, DETR, Poolformer-2022}. ViT \cite{ViT} first proposes a vision transformer using patch embedding as input for image classification, while DETR \cite{DETR} first introduces a transformer encoder-decoder architecture for object detection. Following this, numerous NLP and CV works concentrate on enhancing the token mixing approach of transformers through relative position encoding \cite{PermuteFormer-2021, RPE-2021}, refining attention maps \cite{EfficientAttention-2021, LineFormer-2021}, etc. Recent works find that MLPs \cite{SPACH2022, gMLP-2021}, when employed as token mixers, still achieve competitive performance. Additionally, poolformer \cite{Poolformer-2022} proposes a general architecture without specifying the token mixer. Inspired by these works, we utilize a simple yet efficient pooling operator within the uni-modal encoder to capture global intra-modal context.

\begin{figure*}[t!]
  \centering
  \includegraphics[width=\linewidth]{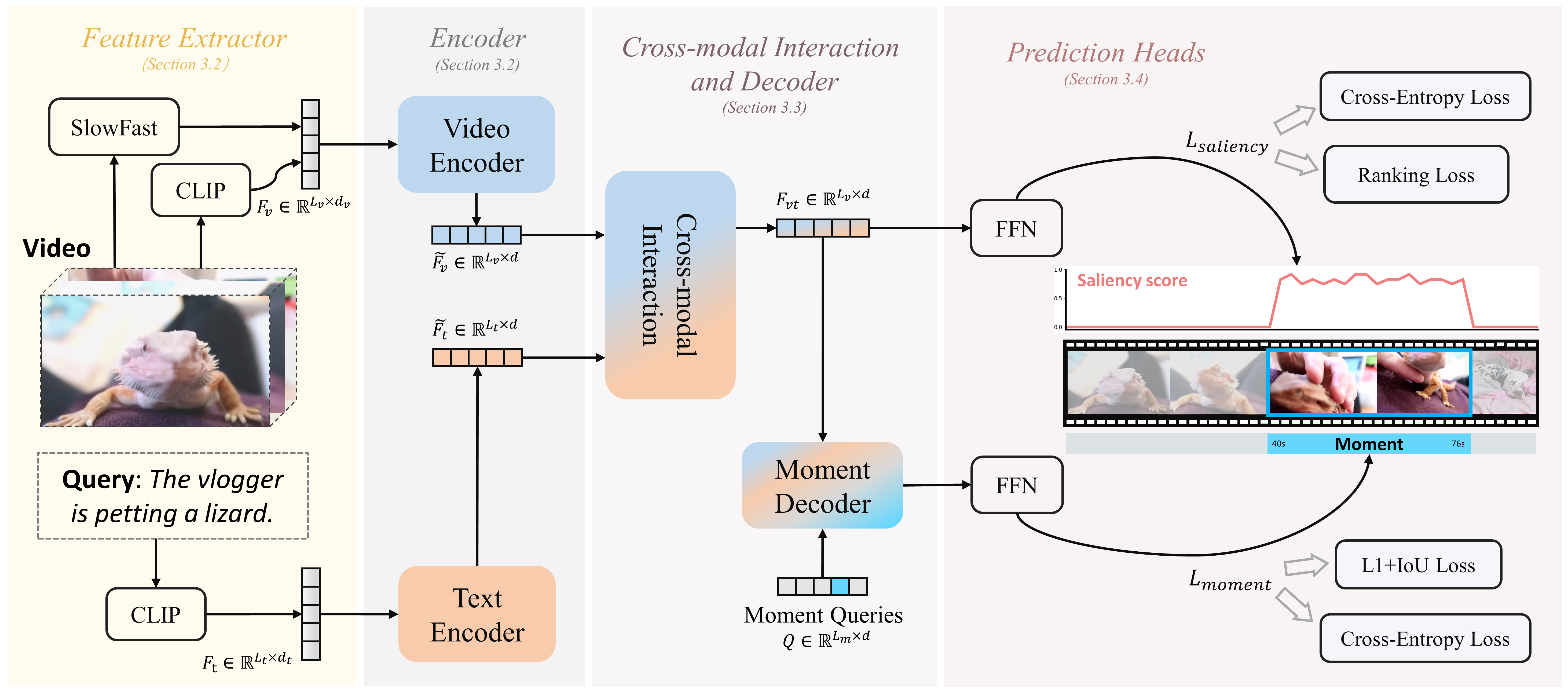}
  \caption{Overall architecture of our model. Given a video and a textual query, we first utilize frozen pretrained models to extract visual and textual features. The encoder (Section \ref{sec:Feature Extractor and Encoder}) models contextualized features under global receptive field. Then, the cross-modal decoder module (Section \ref{sec:Cross-modal Decoder}) fuses features from different modalities. Finally, the prediction heads (Section \ref{sec:Prediction Heads}) generate moment and highlight results, optimized by the losses shown in the above figure. A more comprehensive overview of our model is provided in Section \ref{sec:overview}.}
  \label{Fig:2}
\end{figure*}
%%%%%%%%%%%%%%%%%%%%%%%%%%%%%%%%%%%%%%%%%%%%%%%%%%%%%%%%%%%%%%%%%%%%%%%%%%%%%%%%%%%%
%%%%%%%%%%%%%%%%%%%%%%%%%%%%%%%%%%%%%% Method %%%%%%%%%%%%%%%%%%%%%%%%%%%%%%%%%%%%%%
%%%%%%%%%%%%%%%%%%%%%%%%%%%%%%%%%%%%%%%%%%%%%%%%%%%%%%%%%%%%%%%%%%%%%%%%%%%%%%%%%%%%
\section{Method}

%%%%%%%%%%%%%%%%%%%%%%%%%%%%%%%%%%%%%% Overview %%%%%%%%%%%%%%%%%%%%%%%%%%%%%%%%%%%%%%
\subsection{Overview}
\label{sec:overview}

We define the MHD (video moment and highlight detection) task as follows. Given a video $V \in \mathbb{R}^{L_{v} \times H \times W \times 3}$ containing  $L_{v}$ clips and a natural language query $T \in \mathbb{R}^{L_{t}}$ containing $L_{t}$ words. The goal of MHD is to localize all moments $M=\{ m_{i} \in \mathbb{R}^{2} \}_{i=1}^{L_{m}}$ (where each moment $m_{i}$ consists of a start clip and an end clip) that are highly relevant to $T$, while predicting clip-wise saliency scores $S \in \mathbb{R}^{L_{v}}$ for the whole video simultaneously.

As shown in Figure \ref{Fig:2}, the overall architecture of our proposed model follows a encoder-decoder structure like Moment-DETR \cite{MomentDETR-2021}. Our model consists of four main parts: 

(1) \textit{Feature Extractor.} 
We firstly use frozen pretrained model extract visual features $F_{v} \in \mathbb{R}^{L_{v} \times d_{v}}$ and textual features $F_{t} \in \mathbb{R}^{L_{t} \times d_{t}}$ from given a raw video and a query.

(2) \textit{Encoder.} By capturing the intra-modal correlations, visual and textual features are respectively fed to their respective encoders in order to obtain the contextualized visual features $\tilde{F}_{v} \in \mathbb{R}^{L_{v} \times d}$ and textual features $\tilde{F}_{t} \in \mathbb{R}^{L_{t} \times d}$. 

(3) \textit{Cross-modal Interaction and Decoder.} The cross-modal interaction module fuses contextualized visual and textual features to get the joint moment and highlight representations $F_{vt} \in \mathbb{R}^{L_{v} \times d}$, which are temporally aligned cross-modal features. Subsequently, we utilize the moment decoder with the learnable moment queries $Q \in \mathbb{R}^{L_{m} \times d}$ to derive moment features $\tilde{Q} \in \mathbb{R}^{L_{m} \times d}$.  

(4) \textit{Prediction Heads.} Finally, the simple linear layer with sigmoid activation function is applied to predict moments $M \in \mathbb{R}^{L_{m} \times 2}$ and saliency scores $S \in \mathbb{R}^{L_{v}}$, respectively.

%%%%%%%%%%%%%%%%%%%%%%%%%%%%%%%%%%%%%% Feature Extractor and Encoder %%%%%%%%%%%%%%%%%%%%%%%%%%%%%%%%%%%%%%
\subsection{Feature Extractor and Encoder}
\label{sec:Feature Extractor and Encoder}
\textbf{Feature Extractor.} 
% The input features to the encoder follow Moment-DETR \cite{MomentDETR-2021} settings. Formally, 
Given an untrimmed video, we first sample the video at a frame rate of $1/{\tau}$ to obtain video clips $V \in \mathbb{R}^{L_{v} \times H \times W \times 3}$. Next, we extract visual features $F_{v} \in \mathbb{R}^{L_{v} \times d_{v}}$ using the SlowFast \cite{Slowfast-2019} and CLIP \cite{CLIP-2021} image encoder. For textual features $F_{t} \in \mathbb{R}^{L_{t} \times d_{t}}$, we employ the CLIP text encoder to extract word-wise features.

\vspace{2pt}
\textbf{Encoder.} 
For moment retrieval, some previous works \cite{MCN-2017, CTRL-2017, I2N-2021} utilize sliding windows (temporal convolution networks) to pre-sample proposal candidates from the input video. This sliding-window method leads to redundant computation and low efficiency, as densely sampling candidates with overlap is essential for achieving high accuracy. Furthermore, a significant limitation of this method is that it primary focus on local temporal information, neglecting global temporal context. In MHD task, a comprehensive understanding of global content is essential for achieving high performance \cite{UMT-2022}. 

Given that transformer \cite{Transformer} is capable of effectively modeling long-range information, a multitude of works \cite{FIAN-2020, UMT-2022, TNT-2021, PVT-2021} emphasize the importance of attention mechanisms and focus on designing various attention-based token mixer components for encoder. However, some recent works \cite{SC-Attention-2020, Poolformer-2022, SPACH2022} show that the main contribution to the success of transformer comes from token mixer components and FFNs. Poolformer \cite{Poolformer-2022} models intra-modal correlations and captures global context using a simple pooling operator as the token mixer component. This operator has no learnable parameters, enables each token to evenly aggregate the features of its adjacent tokens. Therefore, we use poolformer as our encoder, which consists of pooling operator and FFN. Additionally, residual connections \cite{ResNet-2016} and layer normalization are applied to each layer of the encoder. Before the encoder, we use separate two-layer FFNs with layer normalization \cite{LayerNormalization-2016} to project visual and textual features into a feature space of the same dimension $d$. For the contextualized visual features $\tilde{F}_{v} \in \mathbb{R}^{L_{v} \times d}$ and textual features $\tilde{F}_{t} \in \mathbb{R}^{L_{t} \times d}$, the encoding process is: 
\begin{equation}
    \bar{F}_{x} = Norm(F_{x} + Pool(F_{x}))
\end{equation}
\begin{equation}
    \tilde{F}_{x} = Norm(\bar{F}_{x} + FFN(\bar{F}_{x}))
\end{equation}
where $F_{x} \in \{ F_{v}, F_{t} \}$. $Norm(\cdot)$ denotes layer normalization. We set the pooling size to $3$ and pooling stride to $1$ for our encoder.

%%%%%%%%%%%%%%%%%%%%%%%%%%%%%%%%%%%%%% Cross-modal Decoder %%%%%%%%%%%%%%%%%%%%%%%%%%%%%%%%%%%%%%
\subsection{Cross-modal Interaction and Decoder}
\label{sec:Cross-modal Decoder}

\textbf{Cross-modal Interaction Module.} 
Most existing works \cite{Survey-VMR-2022, Survey-TSGV-2022} consider cross-modal feature interaction as an essential module, which fuses features from different modalities. The quality of these fused cross-modal features largely determines the performance of moment retrieval. For highlight detection, only a few works \cite{JVA-2021, LPD-2022, VHGNN-2020} focus on cross-modal interaction, since text from the dataset is rarely input into the model. Nonetheless, these works have achieved commendable results. Moreover, since highlight detection requires the clip-level saliency scores, it is necessary to temporally align cross-modal features with visual features. In summary, to accommodate both moment retrieval and highlight detection tasks, we employ the cross-modal interaction module to fuse visual and textual features by emphasizing the portions of visual content most relevant to textual semantics.

\begin{figure}[t]
  \centering
  \includegraphics[width=\linewidth]{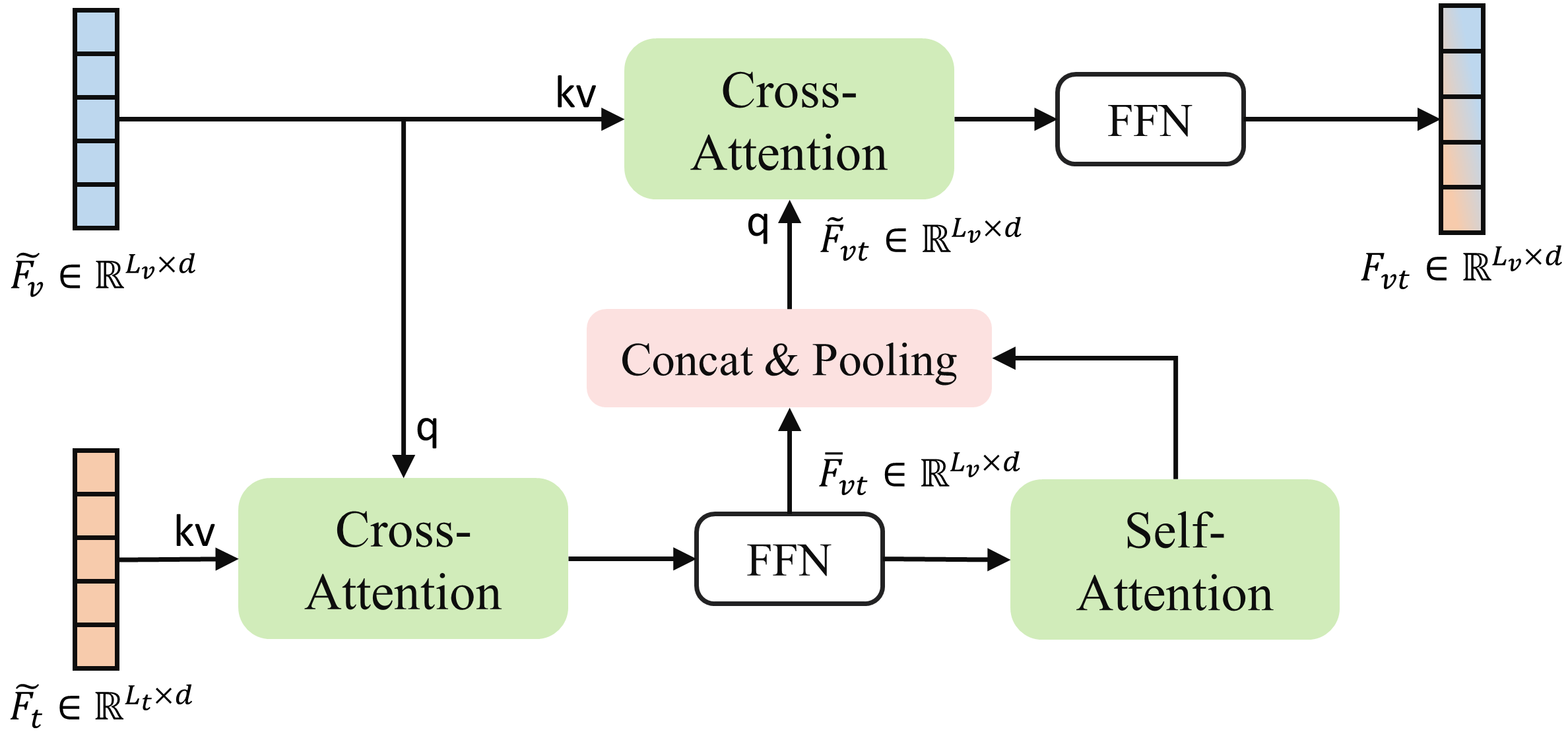}
  \caption{The architecture of the cross-modal interaction module.}
  \label{Fig:3}
\end{figure}

The architecture of the cross-attention module is shown in Figure \ref{Fig:3}. Specifically, we first utilize a cross-attention layer and FFN to dynamically create temporal aligned query features $\bar{F}_{vt} \in \mathbb{R}^{L_{v} \times d}$ derived from textual features. Here, visual features $\tilde{F}_{v}$ as \textit{query}, textual features $\tilde{F}_{t}$ as \textit{key} and \textit{value}. Using the cross-attention layer to calculate the attention weights between video clips and text words, we suppose that each clip can learn which concepts from the text words are present within it. Subsequently, a self-attention layer is applied to aggregate contextualized global query features. We concatenate local and global query features, then use the average pooling operation to get aggregated query features $\tilde{F}_{vt} \in \mathbb{R}^{L_{v} \times d}$. This process can be defined as follows:
\begin{equation}
    \tilde{F} _{vt} = Pool(cat(\bar{F}_{vt}, SA(\bar{F}_{vt})))
\end{equation}
where $SA$ means self-attention. Finally, we use a cross-attention to get the joint moment and highlight representations $F_{vt} \in \mathbb{R}^{L_{v} \times d}$, where aggregated query features $\tilde{F}_{vt}$ as \textit{query} and visual features $\tilde{F}_{v}$ as \textit{key} and \textit{value}. Note that residual connections and layer normalization are applied to all layers. Learnable position encodings \cite{ViT, Transformer} are added to the input of each attention layer.

\vspace{2pt}
\textbf{Moment Decoder.} 
After obtaining the joint moment and highlight representations $F_{vt}$, we follow the method used in existing works \cite{DETR, GTR-2021, MomentDETR-2021} and stack $D_{dec}$ transformer decoder layers as the moment decoder. The moment decoder takes $F_{vt}$ and learnable moment queries $Q \in \mathbb{R}^{L_{m} \times d}$ as inputs, and outputs moment features $\tilde{Q} \in \mathbb{R}^{L_{m} \times d}$.

%%%%%%%%%%%%%%%%%%%%%%%%%%%%%%%%%%%%%%%%%%%%%%%%%%%%%%%%%%%%%%%%%%%%%%%%%%%%%%%%%%%%%%%%%%%%%%%%%%%%%%
%%%%%%%%%%%%%%%%%%%%%%%%%%%%%%%%%%%%%% Table1: QVHighlights test %%%%%%%%%%%%%%%%%%%%%%%%%%%%%%%%%%%%%%
%%%%%%%%%%%%%%%%%%%%%%%%%%%%%%%%%%%%%%%%%%%%%%%%%%%%%%%%%%%%%%%%%%%%%%%%%%%%%%%%%%%%%%%%%%%%%%%%%%%%%%
\begin{table*}[t]
% \captionsetup{justification=centering}
\caption{Performance comparison on QVHighlights \textit{test} split. Results from other models are reported based on existing papers. All models only use visual and textual features and are trained from scratch. The best scores are in \textbf{bold}.}
\label{tab: QVHighlights test}
\begin{adjustbox}{max width=\linewidth}
    \centering
    \begin{tabular}{llcccccccc}
\toprule
\multicolumn{1}{c}{\multirow{3}{*}{Methods}} & \multicolumn{5}{c}{Moment Retrieval}                                               & \multicolumn{2}{c}{Highlight Detection} & \multirow{3}{*}{Params} & \multirow{3}{*}{GFLOPs} \\

\multicolumn{1}{c}{}                         & \multicolumn{2}{c}{R1}          & \multicolumn{3}{c}{mAP}                          & \multicolumn{2}{c}{$\geq$ Very Good}    &                         &                         \\ 
\cmidrule(lr){2-3}
\cmidrule(lr){4-6}
\cmidrule(lr){7-8}
\multicolumn{1}{c}{}                         & @0.5           & @0.7           & @0.5           & @0.75          & Avg.           & mAP                & HIT@1              &                         &                         \\ 
\midrule
BeautyThumb \cite{BeautyThumb-2016}                                  & -              & -              & -              & -              & -              & 14.36              & 20.88              & -                       & -                       \\
DVSE \cite{DVSE-2015}                                         & -              & -              & -              & -              & -              & 18.75              & 21.79              & -                       & -                       \\
MCN \cite{MCN-2017}                                          & 11.41          & 2.72           & 24.94          & 8.22           & 10.67          & -                  & -                  & -                       & -                       \\
CAL \cite{CAL-2019}                                          & 25.49          & 11.54          & 23.40          & 7.65           & 9.89           & -                  & -                  & -                       & -                       \\
XML \cite{TVR-dataset-2020}                                          & 41.83          & 30.35          & 44.63          & 31.73          & 32.14          & 34.49              & 55.25              & -                       & -                       \\
XML+ \cite{TVR-dataset-2020}                                        & 46.69          & 33.46          & 47.89          & 34.67          & 34.90          & 35.38              & 55.06              & -                       & -                       \\ 
\midrule
Moment-DETR \cite{MomentDETR-2021}                                 & 52.89          & 33.02          & 54.82          & 29.40          & 30.73          & 35.69              & 55.60               & \textbf{4.8M}           & \textbf{0.28}           \\
UMT \cite{UMT-2022}                                          & 56.23          & 41.18          & 53.83          & 37.01          & 36.12          & 38.18              & 59.99              & 14.9M                   & 0.63                    \\
\textbf{MH-DETR} (Ours)     & \textbf{60.05} & \textbf{42.48} & \textbf{60.75} & \textbf{38.13} & \textbf{38.38} & \textbf{38.22}     & \textbf{60.51}     & 8.2M                    & 0.34                    \\ 
\bottomrule
    \end{tabular}
\end{adjustbox}
\end{table*}

%%%%%%%%%%%%%%%%%%%%%%%%%%%%%%%%%%%%%%%%%%%%%%%%%%%%%%%%%%%%%%%%%%%%%%%%%%%%%%%%%%%%%%%%%%%%%%%%%%%%%%
%%%%%%%%%%%%%%%%%%%%%%%%%%%%%%%%%%%%%% Table2: QVHighlights val %%%%%%%%%%%%%%%%%%%%%%%%%%%%%%%%%%%%%%%
%%%%%%%%%%%%%%%%%%%%%%%%%%%%%%%%%%%%%%%%%%%%%%%%%%%%%%%%%%%%%%%%%%%%%%%%%%%%%%%%%%%%%%%%%%%%%%%%%%%%%%
\begin{table*}[t]
\caption{Experimental results on QVHighlights \textit{val} split.}
\label{tab: QVHighlights val}
\centering
\begin{adjustbox}{max width=\linewidth}
    \begin{tabular}{@{}lccccccc@{}}
        \toprule
        \multicolumn{1}{c}{Methods} & \multicolumn{5}{c}{Moment Retrieval} & \multicolumn{2}{c}{Highlight Detection} \\ 
         & \multicolumn{2}{c}{R1} & \multicolumn{3}{c}{mAP} & \multicolumn{2}{c}{$\geq$ Very Good} \\
            \cmidrule(lr){2-3}
            \cmidrule(lr){4-6}
            \cmidrule(lr){7-8}
         & R1@0.5 & R1@0.7 & @0.5 & @0.75 & Avg. & mAP & HIT@1 \\ \midrule
        Moment-DETR \cite{MomentDETR-2021} & 53.94 & 34.84 & - & - & 32.20 & 35.65 & 55.55 \\
        UMT \cite{UMT-2022} & - & - & - & - & 37.79 & \textbf{38.97} & - \\
        \textbf{MH-DETR} (Ours) & \textbf{60.84} & \textbf{44.90} & \textbf{60.76} & \textbf{39.64} & \textbf{39.26} & 38.77 & \textbf{61.74} \\ \bottomrule
    \end{tabular}%
\end{adjustbox}
% }
\end{table*}

%%%%%%%%%%%%%%%%%%%%%%%%%%%%%%%%%%%%%% Prediction Heads %%%%%%%%%%%%%%%%%%%%%%%%%%%%%%%%%%%%%%
\subsection{Prediction Heads and Training Loss}
\label{sec:Prediction Heads}

\textbf{Prediction Heads.}
For the joint moment and highlight representations $F_{vt}$, a simple linear layer is applied to predict saliency scores $S \in \mathbb{R}^{L_{v}}$. For the moment features $\tilde{Q}$, we use a linear layer with sigmoid to predict normalized moment start and end points $M \in \mathbb{R}^{L_{m} \times 2}$. Additionally, we use another linear layer with softmax to get class labels. In the MHD task, we set a predicted moment as a \textit{foreground} label if it matches the ground truth, and as a \textit{background} label otherwise.

\vspace{2pt}
\textbf{Saliency Loss.}
The saliency loss $\mathcal{L}_{saliency}$ consists of two components: the weighted binary cross-entropy loss $\mathcal{L}_{bce}$ and the ranking loss $\mathcal{L}_{rank}$. We denote $\lambda_{*}$ are hyperparameters balancing the losses and $\mathcal{L}_{saliency}$ is defined as follows:
\begin{equation}
\mathcal{L}_{saliency} = \lambda_{bce} \mathcal{L}_{bce} + \lambda_{rank} \mathcal{L}_{rank}
\end{equation}
$\mathcal{L}_{bce}$ aims to optimize the saliency score of each clip, which is calculated as:
\begin{equation}
    \mathcal{L}_{bce} = -\sum_{i=1}^{L_{v}} \left[ w_{s} y_i \log \left( s_i \right) + \left( 1-y_i \right) \log \left( 1-s_i \right) \right]
\end{equation}
where $y_{i} \in \{ 0, 1 \}$ and $s_{i}$ represent the saliency ground truth label and predicted saliency score of the \textit{i}-th clip, respectively. Additionally, higher-scoring positive clips carry a higher weight $w_s$ than lower-scoring negative clips.

Similar to previous works \cite{LIM-S-2019, VHGNN-2020, MomentDETR-2021} using contrastive learning strategy, we adopt $\mathcal{L}_{rank}$ to leverage the relationships between two pairs of positive and negative clips, focusing on hard clips. The first pair consists of a highest-scoring clip $s_{high}$ and a lowest-scoring clip $s_{low}$ within the ground truth moments. The second pair includes one clip within and one outside the ground truth moments, with their scores represented as $s_{in}$ and $s_{out}$, respectively. We denote the margin as $\triangle$, and $\mathcal{L}_{rank}$ is defined as:
\begin{equation}
    \mathcal{L}_{rank} = \mathrm{max}(0, \triangle + s_{low} - s_{high}) + \mathrm{max}(0, \triangle + s_{out} - s_{in})
\end{equation}

\vspace{2pt}
\textbf{Moment Loss.}
Since the prediction and ground truth moments do not have a one-to-one correspondence, we follow previous works \cite{GTR-2021, MomentDETR-2021} and employ the Hungarian algorithm to establish the optimal bipartite matching between ground truth and predictions. Assuming there are $L_{n}$ matched ground truth and prediction pairs in a video, we use the span loss $\mathcal{L}_{span}$ to measure the discrepancy between the matched prediction moment $\hat{m}$ and ground truth moment $m$. The span loss is composed of the L1 loss and the generalized IoU loss \cite{GIoU-2019}:
\begin{equation}
    \mathcal{L}_{span} = \sum_{i=1}^{L_{n}} \left[ \lambda_{L1}|| m - \hat{m}||_{1} + \lambda_{IoU} \mathcal{L}_{IoU}(m, \hat{m}) \right]
\end{equation}

Additionally, we employ the weighted binary cross-entropy loss $L_{cls}$ to classify the predicted moments as either foreground or background, which can be formulated as:
\begin{equation}
    \mathcal{L}_{cls} = - \sum_{i=1}^{L_{m}}  \left[ w_{p} z_{i} \log(p_{i}) + (1 - z_{i}) \log(1 - p_{i}) \right]
\end{equation}
where $p_{i}$ and $z_{i}$ represent the predicted probability of the foreground and its corresponding label, respectively. $L_{m}$ is the number of moment queries. The foreground label is assigned a higher weight $w_{p}$ to mitigate label imbalance. Finally, the moment loss $L_{moment}$ is formulated as:
\begin{equation}
    \mathcal{L}_{moment} = \mathcal{L}_{span} + \lambda_{cls} \mathcal{L}_{cls}
\end{equation}

\vspace{2pt}
\textbf{Total Loss.}
The total loss is defined as a linear summation of the losses presented above:
\begin{equation}
    \mathcal{L}_{total} = \mathcal{L}_{saliency} + \mathcal{L}_{moment}
\end{equation}

\section{Experiments}

%%%%%%%%%%%%%%%%%%%%%%%%%%%%%%%%%%%%%% Datasets %%%%%%%%%%%%%%%%%%%%%%%%%%%%%%%%%%%%%%
\subsection{Datasets}
We evaluate the effectiveness of our proposed model on four publicly accessible datasets: QVHighlights \cite{MomentDETR-2021}, Charades-STA \cite{Charades-STA-dataset-2017}, ActivityNet Captions \cite{ActivityNet-Captions-dataset-2017}, and TVSum \cite{TVSum-dataset-2015}.

\vspace{2pt}
\textbf{QVHighlights.} 
QVHighlights is the only dataset presently available for the MHD task. This dataset contains 10,148 diverse YouTube videos with a maximum length of 150 seconds. It contains 10,310 annotations, each including a free-form query, one or more relevant moments (average of about 1.8 moments per query), and clip-wise saliency scores. Moreover, it offers a fair benchmark, as we can only get evaluations by submitting testing split predictions to the QVHighlights online server\footnote{https://codalab.lisn.upsaclay.fr/competitions/6937}. We follow the original QVHighlights data splits on 0.7/0.15/0.15 for training/validation/testing split.

\vspace{2pt}
\textbf{Charades-STA and ActivityNet Captions.}  
Charades-STA and ActivityNet Captions serve as benchmark datasets for moment retrieval. Charades-STA is derived from the original Charades \cite{Charades-dataset-2016} dataset, containing 9,848 videos of daily indoor activities and 16,128 annotations. The standard split of 12,408 and 3,720 annotations is used for training and testing, respectively. ActivityNet Captions is constructed based on the original ActivityNet \cite{ActivityNet-dataset-2015} dataset, containing 19,994 YouTube videos from various domains. As the testing split is reserved for competition, we follow the 2D-TAN \cite{2D-TAN-2020} setting, using 37,421, 17,505, and 17,031 annotations for training, validation, and testing, respectively.

\vspace{2pt}
\textbf{TVSum.}  
TVSum is a benchmark dataset for highlight detection. It contains 10 categories of videos, with 5 videos in each category. In line with UMT \cite{UMT-2022}, we use 0.8/0.2 of the dataset for training and testing.

%%%%%%%%%%%%%%%%%%%%%%%%%%%%%%%%%%%%%% Evaluation Metrics %%%%%%%%%%%%%%%%%%%%%%%%%%%%%%%%%%%%%%
\subsection{Evaluation Metrics}
We use the same evaluation metrics following existing works. Specifically, for QVHighlights, we use Recall@1 with thresholds 0.5 and 0.7, mean average precision (mAP) with IoU thresholds 0.5 and 0.75, and the average mAP over multiple IoU thresholds [0.5:0.05:0.95] for moment retrieval. For highlight detection, mAP and HIT@1 are used, where HIT@1 is used to compute the hit ratio for the highest-scored clip. We also report parameters and GFLOPs (input visual features  $\in \mathbb{R}^{75 \times 2816}$  and textual features $\in \mathbb{R}^{32\times 512}$). For Charades-STA and ActivityNet Captions, Recall@1 with IoU thresholds 0.5 and 0.7 are used. For TVSum, the mAP at top-5 serves as the metric.

%%%%%%%%%%%%%%%%%%%%%%%%%%%%%%%%%%%%%% Experimental Settings %%%%%%%%%%%%%%%%%%%%%%%%%%%%%%%%%%%%%%
\subsection{Experimental Settings}
The setting of the pretrained feature extractor for QVHighlights is detailed in Section \ref{sec:Feature Extractor and Encoder}. For Charades-STA, we use official VGG \cite{VGG-2016} features and GloVe \cite{GloVe-2014} textual embeddings. To enable more comprehensive comparisons, we also utilize I3D \cite{I3D-2017} visual features provided by MIGCN \cite{MIGCN-2021}. For ActivityNet Captions, we use C3D \cite{C3D-2015} visual features and GloVe textual embeddings. For TVSum, we following UMT use I3D features pretrained on Kinetics-400 \cite{Kinetics-400-dataset-2017} as visual features and CLIP features as textual features.

For input raw video, we set the sampling rate $1/\tau$ to $1/8$. The hidden dimension is set to $d=256$, and the number of moment queries $L_{m} = 10$.  The maximum input query length $L_{t}$ is set to 32, 10, and 50 for QVHighlights, Charades-STA, and ActivityNet Captions, respectively. The number of layers in the encoder, cross-modal interaction module, and moment decoder is configured as 1/1/4. We use dropout of 0.1 and drop-path \cite{DropPath-2016} of 0.1 for all multi-head attention layers and FFNs. Additionally, we adopt an extra dropout with rates of 0.5 and 0.3 for visual and textual projection layers, respectively. The post-norm style layer normalization and ReLU \cite{ReLU-2013} activation are used in the model. The hyperparameters of losses are set as follows: $\lambda_{bce}=1$, $\lambda_{rank}=0.1$, $\lambda_{L1}=10$, $\lambda_{IoU}=1$, $\lambda_{cls}=4$, $w_{s}=5$, $w_{p}=10$, $\triangle=0.2$. In all experiments, we use AdamW \cite{AdamW-2017} optimizer with 2e-4 learning rate and 1e-4 weight decay. The model is trained with batch size 32 for 200 epochs on QVHighlights, batch size 8 (32) for 100 epochs on Charades-STA with VGG (I3D) features, batch size 32 for 100 epochs on ActivityNet Captions, and batch size 2 for 1000 epochs on TVSum, respectively. All experiments are conducted using a single GeForce RTX 2080Ti GPU. More details are available in our released code.

%%%%%%%%%%%%%%%%%%%%%%%%%%%%%%%%%%%%%%%%%%%%%%%%%%%%%%%%%%%%%%%%%%%%%%%%%%%%%%%%%%%%%%%%%%%%%%%%%%%%%%
%%%%%%%%%%%%%%%%%%%%%%%%%%%%%%%%%%%%%% Table3: Charades-STA test %%%%%%%%%%%%%%%%%%%%%%%%%%%%%%%%%%%%%%
%%%%%%%%%%%%%%%%%%%%%%%%%%%%%%%%%%%%%%%%%%%%%%%%%%%%%%%%%%%%%%%%%%%%%%%%%%%%%%%%%%%%%%%%%%%%%%%%%%%%%%
\begin{table}[t!]
% \captionsetup{justification=centering}
\caption{Comparison with representative moment retrieval models on Charades-STA \textit{test} split. All models use either the official VGG features or the I3D features.}
\label{tab: Charades-STA test}
\centering
\begin{adjustbox}{max width=\linewidth}
% \resizebox{0.75\linewidth}{!}{%
\begin{tabular}{@{}clcc@{}}
    \toprule
    Feature              & \multicolumn{1}{c}{Methods} & R1@0.5 & R1@0.7 \\ \midrule
    \multirow{6}{*}{VGG} & SAP \cite{SAP-2019}                        & 27.42  & 13.36  \\
                         & 2D-TAN \cite{2D-TAN-2020}                     & 40.94  & 22.85  \\
                         & CMAS \cite{CMAS-2022}                       & 48.37  & 29.44  \\
                         & UMT \cite{UMT-2022}                        & 49.35  & 26.16  \\
                         & Moment-DETR \cite{MomentDETR-2021}                 & 53.63  & 31.37  \\
                         & \textbf{MH-DETR} (Ours)     & \textbf{55.47} & \textbf{32.41}  \\ \midrule
    \multirow{6}{*}{I3D} & MAN \cite{MAN-2019}                        & 46.63  & 22.72  \\
                         & DRN \cite{DRN-2020}                        & 53.09  & 31.75  \\
                         & SCDM \cite{SCDM-2019}                       & 54.44  & 33.43  \\
                         & VSLNet \cite{VSLNet-2020}                     & 54.19  & 35.22  \\
                         & MIGCN \cite{MIGCN-2021}                      & \textbf{57.10}  & 34.54  \\
                         & \textbf{MH-DETR} (Ours)     & 56.37  & \textbf{35.94}  \\ \bottomrule
    \end{tabular}%
% }
\end{adjustbox}
\end{table}

%%%%%%%%%%%%%%%%%%%%%%%%%%%%%%%%%%%%%%%%%%%%%%%%%%%%%%%%%%%%%%%%%%%%%%%%%%%%%%%%%%%%%%%%%%%%%%%%%%%%%%
%%%%%%%%%%%%%%%%%%%%%%%%%%%%%%%%%%%%%%%%%%% Table 4 %%%%%%%%%%%%%%%%%%%%%%%%%%%%%%%%%%%%%%%%%%%%%%%%
%%%%%%%%%%%%%%%%%%%%%%%%%%%%%%%%%%%%%%%%%%%%%%%%%%%%%%%%%%%%%%%%%%%%%%%%%%%%%%%%%%%%%%%%%%%%%%%%%%%%%%
\begin{table}[t]
\centering
\caption{Experimental results on ActivityNet Captions \textit{test} split.}
\label{tab: ActivityNet Captions test}
    % \resizebox{0.57\linewidth}{!}{%
        \begin{tabular}{@{}lcc@{}}
            \toprule
            \multicolumn{1}{c}{Methods} & R1@0.5 & R1@0.7 \\ \midrule
            QSPN \cite{QSPN-2019}                       & 27.70  & 13.60  \\
            2D-TAN \cite{2D-TAN-2020}                     & 44.51  & 26.54  \\
            MIGCN \cite{MIGCN-2021}                       & 44.94  & -      \\
            HiSA \cite{HiSA-2022}                      & 45.36  & 27.68  \\
            CMAS \cite{CMAS-2022}                       & 46.23  & 29.48  \\
            VISA \cite{VISA-2022}                       & 47.13  & 29.64  \\
            \textbf{MH-DETR} (Ours)     & \textbf{47.15}  & \textbf{30.86} \\ \bottomrule
        \end{tabular}%
    % }
\end{table}

%%%%%%%%%%%%%%%%%%%%%%%%%%%%%%%%%%%%%% Experimental Results %%%%%%%%%%%%%%%%%%%%%%%%%%%%%%%%%%%%%%
\subsection{Experimental Results}

%%%%%%%%%%%%%%%%%%%%%%%%%%%%%%%%%%%%%%%%%%%%%%%%%%%%%%%%%%%%%%%%%%%%%%%%%%%%%%%%%%%%%%%%%%%%%%%%%%%%%%
%%%%%%%%%%%%%%%%%%%%%%%%%%%%%%%%%%%%%%%%%%% Table 5 %%%%%%%%%%%%%%%%%%%%%%%%%%%%%%%%%%%%%%%%%%%%%%%%
%%%%%%%%%%%%%%%%%%%%%%%%%%%%%%%%%%%%%%%%%%%%%%%%%%%%%%%%%%%%%%%%%%%%%%%%%%%%%%%%%%%%%%%%%%%%%%%%%%%%%%

\begin{table*}[ht]
\centering
    \caption{Performance comparison with representative highlight detection models on TVSum dataset. The metric is top-5 mAP.}
    \label{tab: TVsum}
        \begin{tabular}{@{}lccccccccccc@{}}
            \toprule
            \multicolumn{1}{c}{Methods} & VT   & VU   & GA   & MS   & PK   & PR   & FM   & BK   & BT   & DS   & Avg. \\ \midrule
            sLSTM \cite{sLSTM-2016}                       & 41.1 & 46.2 & 46.3 & 47.7 & 44.8 & 46.1 & 45.2 & 40.6 & 47.1 & 45.5 & 45.1 \\
            SG \cite{SG-2017}                         & 42.3 & 47.2 & 47.5 & 48.9 & 45.6 & 47.3 & 46.4 & 41.7 & 48.3 & 46.6 & 46.2 \\
            LIM-S \cite{LIM-S-2019}                      & 55.9 & 42.9 & 61.2 & 54.0 & 60.4 & 47.5 & 43.2 & 66.3 & 69.1 & 62.6 & 56.3 \\
            Trailer \cite{Trailer-2020}                     & 61.3 & 54.6 & 65.7 & 60.8 & 59.1 & 70.1 & 58.2 & 64.7 & 65.6 & 68.1 & 62.8 \\
            SL-Module \cite{SL-Module-2021}                   & \textbf{86.5} & 68.7 & 74.9 & \textbf{86.2} & 79.0 & 63.2 & 58.9 & 72.6 & 78.9 & 64.0 & 73.3 \\
            \textbf{MH-DETR} (Ours)     & 86.1 & \textbf{79.4} & \textbf{84.3} & 85.8 & \textbf{81.2} & \textbf{83.9} & \textbf{74.3} & \textbf{82.7} & \textbf{86.5} & \textbf{71.6} & \textbf{81.6} \\ \bottomrule
        \end{tabular}%
\end{table*}

\begin{figure*}[t]
  \centering
  \includegraphics[width=\linewidth]{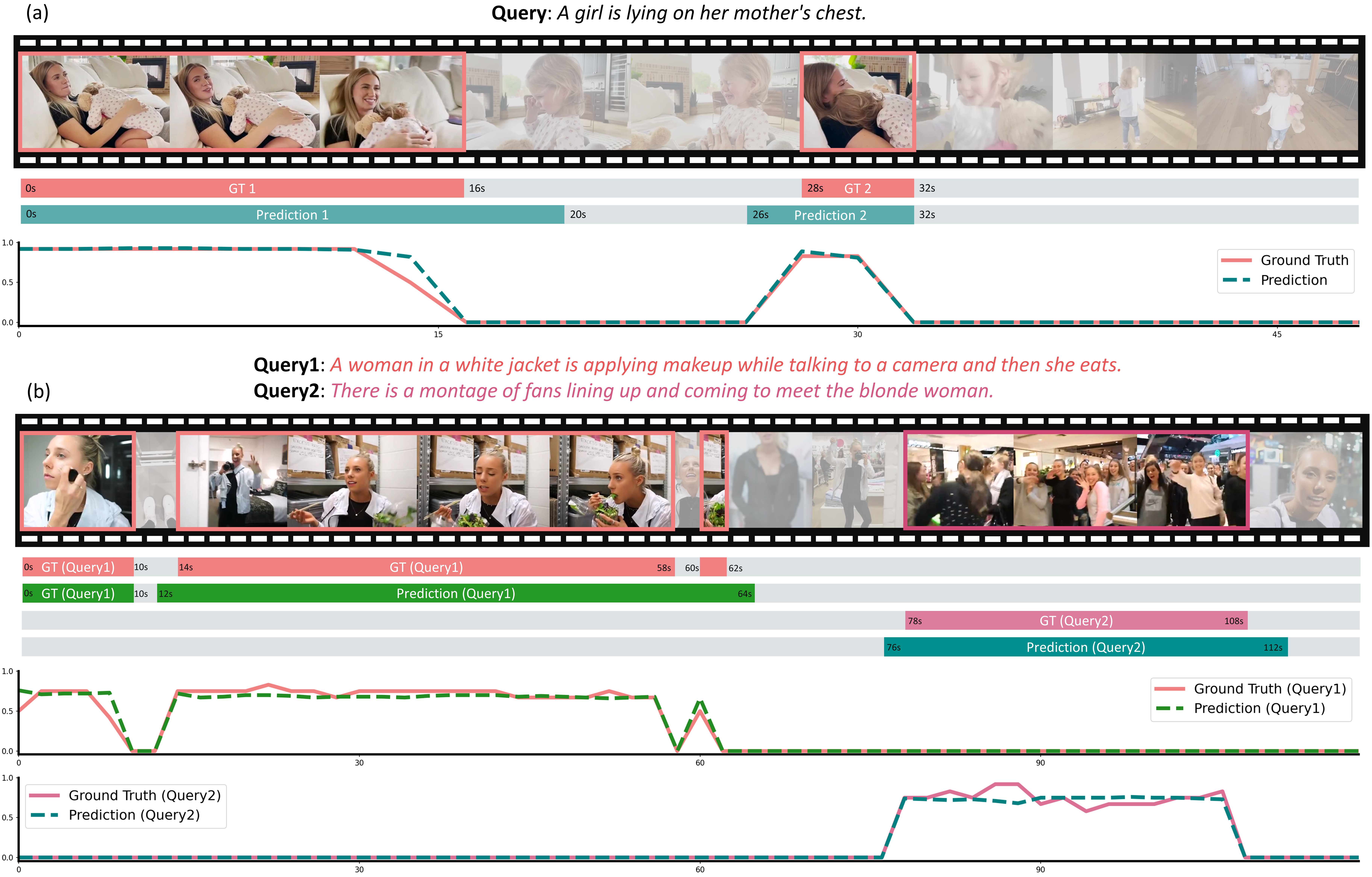}
  \caption{Prediction visualization on QVHighlights \textit{val} split. Images displayed from top to bottom present the input query and video, as well as the predicted moments and highlights. (a) Our model can effectively predict multiple moments and highlights that correspond to the same query. (b) Our model can also handle situations where multiple queries are present in the same video.}
  \label{Fig:4}
\end{figure*}

\begin{table*}[t]
\caption{Effectiveness of each module in our proposed MH-DETR on QVHighlights \textit{val} split, where "enc.", "crs-int.", and "mom-dec." denote the encoder in Section \ref{sec:Feature Extractor and Encoder}, the cross-modal interaction module, and the moment decoder in Section \ref{sec:Cross-modal Decoder}, respectively. MR and HD represent moment retrieval and highlight detection respectively. VG is the abbreviation for very good.}
\label{tab: ablations on modules}
\centering
\begin{adjustbox}{max width=\linewidth}
\begin{tabular}{@{}ccccccccccc@{}}
    \toprule
    \multirow{2}{*}{Model} & \multicolumn{3}{c}{Modules} & \multicolumn{5}{c}{MR} & \multicolumn{2}{c}{HD ($\geq$VG)} \\ 
        \cmidrule(lr){2-4}
        \cmidrule(lr){5-9}
        \cmidrule(lr){10-11}
     & enc. & crs-int. & mom-dec. & R1@0.5 & R1@0.7 & mAP@0.5 & mAP@0.75 & mAP Avg. & mAP & HIT@1 \\ \midrule
    \ding{172} & $\boldsymbol{\checkmark}$ &  &  & 41.87 & 20.19 & 45.32 & 18.63 & 21.21 & 32.85 & 54.71 \\
    \ding{173} & $\boldsymbol{\checkmark}$ & $\boldsymbol{\checkmark}$ &  & 56.32 & 36.58 & 56.02 & 31.36 & 32.63 & 37.48 & 59.84 \\
    \ding{174} & $\boldsymbol{\checkmark}$ &  & $\boldsymbol{\checkmark}$ & 57.16 & 38.90 & 58.42 & 34.98 & 35.04 & 34.14 & 56.90 \\
    \ding{175} &  & $\boldsymbol{\checkmark}$ & $\boldsymbol{\checkmark}$ & \textbf{61.10} & 43.68 & 60.21 & 37.83 & 37.89 & 38.71 & 61.72 \\
    \ding{176} & $\boldsymbol{\checkmark}$ & $\boldsymbol{\checkmark}$ & $\boldsymbol{\checkmark}$ & 60.84 & \textbf{44.90} & \textbf{60.76} & \textbf{39.64} & \textbf{39.26} & \textbf{38.77} & \textbf{61.74} \\ \bottomrule
\end{tabular}%
\end{adjustbox}
\end{table*}

\begin{table*}[t]
\caption{Ablation study of losses on QVHighlights \textit{val} split.}
\label{tab: ablations on losses}
\centering
\begin{adjustbox}{max width=\linewidth}
\begin{tabular}{@{}ccccccccccccc@{}}
    \toprule
    \multirow{2}{*}{Index} & \multicolumn{5}{c}{Losses} & \multicolumn{5}{c}{MR} & \multicolumn{2}{c}{HD ($\geq$VG)} \\ 
        \cmidrule(lr){2-6}
        \cmidrule(lr){7-11}
        \cmidrule(lr){12-13}
     & $\mathcal{L}_{cls}$ & $\mathcal{L}_{L1}$ & $\mathcal{L}_{IoU}$ & $\mathcal{L}_{bce}$ & $\mathcal{L}_{rank}$ & R1@0.5 & R1@0.7 & mAP@0.5 & mAP@0.75 & mAP Avg. & mAP & HIT@1 \\ \midrule
    (1) &  &  &  & $\boldsymbol{\checkmark}$ & $\boldsymbol{\checkmark}$ & - & - & - & - & - & 35.04 & 53.81 \\
    (2) & $\boldsymbol{\checkmark}$ & $\boldsymbol{\checkmark}$ & $\boldsymbol{\checkmark}$ &  &  & 54.84 & 37.10 & 57.26 & 33.42 & 33.50 & - & - \\
    (3) & $\boldsymbol{\checkmark}$ &  & $\boldsymbol{\checkmark}$ & $\boldsymbol{\checkmark}$ & $\boldsymbol{\checkmark}$ & 55.87 & 38.06 & 56.08 & 33.90 & 34.02 & 37.83 & 61.29 \\
    (4) & $\boldsymbol{\checkmark}$ & $\boldsymbol{\checkmark}$ &  & $\boldsymbol{\checkmark}$ & $\boldsymbol{\checkmark}$ & 57.61 & 40.00 & 55.43 & 34.33 & 34.38 & 38.62 & 60.71 \\
    (5) & $\boldsymbol{\checkmark}$ & $\boldsymbol{\checkmark}$ & $\boldsymbol{\checkmark}$ &  & $\boldsymbol{\checkmark}$ & 56.71 & 38.71 & 57.70 & 34.33 & 35.30 & 36.92 & 56.13 \\
    (6) & $\boldsymbol{\checkmark}$ & $\boldsymbol{\checkmark}$ & $\boldsymbol{\checkmark}$ & $\boldsymbol{\checkmark}$ &  & 60.45 & 44.52 & 58.02 & 37.20 & 37.26 & 38.22 & 60.19 \\
    (7) & $\boldsymbol{\checkmark}$ & $\boldsymbol{\checkmark}$ & $\boldsymbol{\checkmark}$ & $\boldsymbol{\checkmark}$ & $\boldsymbol{\checkmark}$ & \textbf{60.84} & \textbf{44.90} & \textbf{60.76} & \textbf{39.64} & \textbf{39.26} & \textbf{38.77} & \textbf{61.74} \\ 
    \bottomrule
\end{tabular}%
\end{adjustbox}
\end{table*}

\begin{table*}[t]
\caption{Comparison with Moment-DETR using the same parameters on QVHighlights \textit{val} split.}
\label{tab: ablations on fair model}
\centering
\begin{adjustbox}{max width=\linewidth}
\begin{tabular}{@{}lccccccccc@{}}
    \toprule
    \multicolumn{1}{c}{\multirow{2}{*}{Methods}} & \multicolumn{5}{c}{MR} & \multicolumn{2}{c}{HD ($\geq$VG)} & \multirow{2}{*}{Params} & \multirow{2}{*}{GFLOPs} \\ 
        \cmidrule(lr){2-6}
        \cmidrule(lr){7-8}
    \multicolumn{1}{c}{} & R1@0.5 & R1@0.7 & mAP@0.5 & mAP@0.75 & mAP Avg. & mAP & HIT@1 &  &  \\ 
        \midrule
    Moment-DETR \cite{MomentDETR-2021} & 53.94 & 34.84 & - & - & 32.20 & 35.65 & 55.55 & \textbf{4.8M} & 0.28 \\
    \textbf{MH-DETR-S} (Ours) & \textbf{60.19} & \textbf{44.52} & \textbf{59.59} & \textbf{38.59} & \textbf{37.77} & \textbf{38.20} & \textbf{59.29} & 5.0M & \textbf{0.24} \\
        \midrule
    Moment-DETR-L & 55.35 & 37.61 & 56.28 & 32.07 & 32.96 & 36.25 & 56.39 & 8.5M & 0.49 \\
    \textbf{MH-DETR} (Ours) & \textbf{60.84} & \textbf{44.90} & \textbf{60.76} & \textbf{39.64} & \textbf{39.26} & \textbf{38.77} & \textbf{61.74} & \textbf{8.2M} & \textbf{0.34} \\
    \bottomrule
\end{tabular}%
\end{adjustbox}
\end{table*}

\textbf{Moment and Highlight Detection.}
We begin by comparing our proposed MH-DETR with existing methods on QVHighlights \textit{test} split. To ensure a fair comparison, all methods only use visual and textual features and are trained from scratch. The results are presented in Table \ref{tab: QVHighlights test}. Our proposed MH-DETR outperforms the state-of-the-art (SOTA) method UMT by 3.82\% and 6.92\% on the R1@0.5 and mAP@0.5 metrics, respectively. Table \ref{tab: QVHighlights val} report the comparison results between our method and existing methods on QVHighlights \textit{val} split. Notably, MH-DETR also outperforms other methods. Additionally, we visualize the predictions of our method on QVHighlights in Figure \ref{Fig:4}. Images displayed from top to bottom present the input query and video, as well as the predicted moments and highlights. Figure \ref{Fig:4} (a) indicates our model can effectively predict multiple moments and highlights that correspond to the same query. Figure \ref{Fig:4} (b) represents  that our model can also handle situations where multiple queries are present in the same video.

\vspace{2pt}
\textbf{Moment Retrieval.}
Table \ref{tab: Charades-STA test} reports the results of our MH-DETR compared with other methods on Charades-STA \textit{test} split. For more comprehensive comparisons, we use two types of visual features: VGG and I3D. Our method achieves the best results in almost all metrics. Specifically, using the same VGG features, MH-DETR outperforms the existing method Moment-DETR by 1.84\% and 1.04\% on the R1@0.5 and R1@0.7 metrics, respectively. Regarding I3D features, MH-DETR demonstrates competitive performance compared to the recent anchor-based method MIGCN. Table \ref{tab: ActivityNet Captions test} shows that our method better than the recent proposal-free SOTA method VISA \cite{VISA-2022}. In line with MIGCN, we use only C3D visual features to ensure fair comparisons.

\vspace{2pt}
\textbf{Highlight Detection.} The results of highlight detection on TVSum are presented in Table \ref{tab: TVsum}, where MH-DETR achieves the best results in almost categories. Specifically, although our method is slightly inferior to the recent SL-Module \cite{SL-Module-2021} in the changing Vehicle Tire (VT) and Making Sandwich (MS), it surpasses SL-Module by 8.3\% in terms of average top-5 mAP across all categories. Additionally, MH-DETR outperforms SL-Module by 15.4\% and 20.7\% in the Flash Mob Gathering (FM) and Grooming an Animal (GA) categories, respectively.

%%%%%%%%%%%%%%%%%%%%%%%%%%%%%%%%%%%%%% Ablation Studies %%%%%%%%%%%%%%%%%%%%%%%%%%%%%%%%%%%%%%
\subsection{Ablation Studies}
\label{subsec: Ablation Studies}

To evaluate the effectiveness of each module in our method, we conduct in-depth ablation studies presented in Table \ref{tab: ablations on modules}. The process of calculating the loss and establishing bipartite matching is the same as described in Section \ref{sec:Prediction Heads}. Model \ding{174} serves as our baseline model, based on Moment-DETR, and replaces transformer with poolformer in the encoder. Model \ding{174} demonstrates a modest performance enhancement compared to Moment-DETR, suggesting that a simple token mixer with a FFN is adequate for the encoder. Model \ding{172}, created by removing the moment decoder from model \ding{174}, directly uses the cross-attention layer in the encoder with visual features as \textit{query} and textual features as \textit{key} and \textit{value}. Model \ding{172} shows a performance drop on MR task, indicating the effectiveness of the moment decoder in capturing moment features. Model \ding{175}, which removes the encoder from the full MH-DETR (model \ding{176}), exhibits only a slight decrease in performance. This suggests that the model has minimal reliance on the uni-modal encoder, and the cross-modal interaction module is also effective in modeling global context. Models \ding{173} and \ding{176} add the cross-modal interaction module, resulting in a significant performance improvement in MHD task compared to models \ding{172} and \ding{174}. This demonstrates that this module effectively fuses visual and textual features, yielding robust joint moment and highlight features. 

Table \ref{tab: ablations on losses} displays the performance of MH-DETR when using different combinations of losses. Rows (3)-(7) investigate the impact of each individual loss. Rows (1), (2), and (7) reveal that turning off either the highlight loss or the moment loss leads to a significant performance drop for both HD and MR tasks. This demonstrates the importance of co-optimization in achieving significant performance improvements for MHD task."

In addition, compared to Moment-DETR, the parameters of MH-DETR have increased, suggesting that our model can fit more robust functions. To ensure a fair comparison, we raise the number of layers in both the encoder and decoder of Moment-DETR to 4, denoted as Moment-DETR-L. We also remove the encoder of MH-DETR and reduced the number of moment decoder layers to 2, denoted as MH-DETR-S. As illustrated in Table \ref{tab: ablations on fair model}, our model achieves the best results with the same parameters.

%%%%%%%%%%%%%%%%%%%%%%%%%%%%%%%%%%%%%%%%%%%%%%%%%%%%%%%%%%%%%%%%%%%%%%%%%%%%%%%%%%%%%%%%
%%%%%%%%%%%%%%%%%%%%%%%%%%%%%%%%%%%%%% Conclusion %%%%%%%%%%%%%%%%%%%%%%%%%%%%%%%%%%%%%%
%%%%%%%%%%%%%%%%%%%%%%%%%%%%%%%%%%%%%%%%%%%%%%%%%%%%%%%%%%%%%%%%%%%%%%%%%%%%%%%%%%%%%%%%
\section{Conclusion and Future Work}
\label{sec: Conclusion}

In this paper, we propose a novel model called \textbf{MH-DETR} (\textbf{M}oment and \textbf{H}ighlight \textbf{DE}tection \textbf{TR}ansformer) for MHD (video moment and highlight detection) task. Specifically, we introduce a simple yet efficient pooling operator within the uni-modal encoder to capture global intra-modal information. Furthermore, we design the cross-modal interaction module that integrates visual and textual features to obtain temporally aligned cross-modal features. Extensive experiments on QVHighlights, Charades-STA, ActivityNet Captions, and TVSum datasets demonstrate that our method outperforms SOTA method, underscoring the effectiveness and superiority of our proposed MH-DETR.

To further explore the potential of cross-modal models in MHD task, several future works are beneficial. One attempt is to introduce more modal features, such as optical flow, depth and object features \cite{DRFT-2021, LEORN-2022, DORi-2021}. Another direction is to integrate advancements from follow-up works of DETR \cite{DAB-DETR-2022, DINO-2022, DN-DETR-2022}. Moreover, developing a loss function that simultaneously considers both MR and HD tasks would be valuable.

%%%%%%%%%%%%%%%%%%%%%%%%%%%%%%%%%%%%%%%%%%%%%%%%%%%%%%%%%%%%%%%%%%%%%%%%%%%%%%%%%%%%%%%%
%%%%%%%%%%%%%%%%%%%%%%%%%%%%%%%%%%%%%% references %%%%%%%%%%%%%%%%%%%%%%%%%%%%%%%%%%%%%%
%%%%%%%%%%%%%%%%%%%%%%%%%%%%%%%%%%%%%%%%%%%%%%%%%%%%%%%%%%%%%%%%%%%%%%%%%%%%%%%%%%%%%%%%
% \clearpage
\bibliographystyle{ACM-Reference-Format}
\bibliography{paper.bib}

%%%%%%%%%%%%%%%%%%%%%%%%%%%%%%%%%%%%%%%%%%%%%%%%%%%%%%%%%%%%%%%%%%%%%%%%%%%%%%%%%%%%%%
%%%%%%%%%%%%%%%%%%%%%%%%%%%%%%%%%%%%%% appendix %%%%%%%%%%%%%%%%%%%%%%%%%%%%%%%%%%%%%%
% \clearpage

% \appendix

% \section{Appendix}


\end{document}